# A Hybrid Model for Few-Shot Text Classification Using Transfer and Meta-Learning

Jia Gao
Stevens Institute of Technology
Hoboken, USA

Shuangquan Lyu
Carnegie Mellon University
Pittsburgh, USA

Guiran Liu
San Francisco State University
San Francisco, USA

Binrong Zhu
San Francisco State University
San Francisco, USA

Hongye Zheng
The Chinese University of Hong Kong
Hong Kong, China

Xiaoxuan Liao*
New York University
New York City, USA

*Abstract*-With the continuous development of natural language processing (NLP) technology, text classification tasks have been widely used in multiple application fields. However, obtaining labeled data is often expensive and difficult, especially in few-shot learning scenarios. To solve this problem, this paper proposes a few-shot text classification model based on transfer learning and meta-learning. The model uses the knowledge of the pre-trained model for transfer and optimizes the model's rapid adaptability in few-sample tasks through a meta-learning mechanism. Through a series of comparative experiments and ablation experiments, we verified the effectiveness of the proposed method. The experimental results show that under the conditions of few samples and medium samples, the model based on transfer learning and meta-learning significantly outperforms traditional machine learning and deep learning methods. In addition, ablation experiments further analyzed the contribution of each component to the model performance and confirmed the key role of transfer learning and meta-learning in improving model accuracy. Finally, this paper discusses future research directions and looks forward to the potential of this method in practical applications.

## I. INTRODUCTION

With the rapid development of natural language processing (NLP) technology, text classification, as one of its core tasks, has achieved significant applications in many fields [1]. From sentiment analysis to spam filtering, from news recommendation to Human-Computer interaction applications [2], the demand for text classification in practical applications is increasing. However, in many real-world scenarios, obtaining large-scale annotated datasets is challenging, particularly in specialized domains or low-resource languages, where annotation is costly and data is scarce [3]. Therefore, training an efficient text classification model with good generalization ability under the condition of a small number of samples has become an important challenge in current NLP research.

To address data scarcity in text classification, various x-shot learning paradigms have been explored, including zero-shot and few-shot learning [4]. Zero-shot learning enables classification without any labeled examples but often relies on external knowledge sources that may not be reliable. One-shot learning allows classification with only one labeled instance per class but struggles with complex tasks where a single example may not represent category variations [5]. Few-shot learning, in contrast, provides a balanced approach by leveraging a small but sufficient number of labeled samples to train effective models. Given its practicality in real-world applications, few-shot learning has become a key research direction in NLP, particularly for text classification. [6].

The goal of few-shot learning is to train a model that can work effectively in a new task with only a small number of annotated samples. Traditional text classification models usually rely on large-scale annotated data for training, but in many practical applications, the acquisition of annotated data is expensive and time-consuming. To meet this challenge, few-shot learning simulates the situation of a small number of samples in the learning task to build a model that can effectively learn under limited data conditions. In the framework of few-shot learning, transfer learning and meta-learning, as two main technical means, have been widely used in text classification tasks and have achieved relatively ideal results.

The basic idea of transfer learning is to transfer the knowledge learned from the source task to the target task, especially when the data in the target task is scarce, transfer learning can significantly reduce the dependence on labeled data [7]. Specifically, by pre-training a large-scale language model [8], such as BERT, GPT, etc., and then migrating it to a specific text classification task, the performance of the model can be effectively improved. Especially when the amount of labeled data for the target task is small, transfer learning can make up for the lack of target task data by sharing the knowledge in the source task. The application of transfer learning in few-shot learning can learn general language knowledge through pre-training on a large amount of unlabeled data, so as to achieve better results in the target task.

At the same time, meta-learning, as a technology that enables the model to quickly adapt to new tasks, also provides new ideas for few-shot learning. The core of meta-learning is to learn how to learn so that the model can quickly adjust parameters and obtain better performance with only a small number of samples when it encounters a new task. In text classification tasks, meta-learning algorithms (such as MAML, ProtoNet, etc.) optimize the learning process of the model in a few-sample scenario, so that the model can quickly capture the essential characteristics of the task when facing different categories and samples, thereby improving the classification effect [9].

When designing and optimizing a few-sample text classification model based on transfer learning, many factors need to be considered. First, how to choose a suitable pre-trained model and transfer strategy is an important issue. Different pre-trained models have different learning abilities for language features, so it is crucial to choose a model that is highly relevant to the target task. Secondly, how to avoid the domain gap between the source task and the target task during transfer learning to make the transfer process smoother is also a difficult problem that needs to be solved in model design. Finally, in the case of a few samples, how to further improve the robustness and generalization ability of the model through technologies such as data enhancement, label smoothing, and knowledge distillation is also a hot topic in current research.

With the continuous advancement of deep learning and NLP technology, the few-sample text classification model based on transfer learning is gradually maturing and has achieved remarkable results in many practical applications. For example, in the medical field, although labeled data is relatively scarce, transfer learning technology can quickly adapt to new disease classification tasks; in social media, facing the ever-changing sentiment classification tasks, the combination of transfer learning and few-shot learning enables the model to adapt to new sentiment trends in real-time. In the future, with the emergence of more pre-trained models and optimization algorithms, few-shot text classification technology based on transfer learning will further expand its application scope and provide more efficient technical support for the intelligent development of various industries.

## II. RELATED WORK

Few-shot learning for text classification requires models that can efficiently generalize from limited labeled data. Recent advancements in transfer learning, meta-learning, and auxiliary optimization techniques provide key foundations for this area, helping address the challenges of data scarcity and rapid task adaptation.

Transfer learning, particularly through large pre-trained models such as BERT and GPT, has shown significant potential in enhancing downstream NLP performance by transferring general language knowledge learned from large corpora to specific tasks. Recent work utilizing transformer-based architectures demonstrates how leveraging pre-trained models can improve privacy-aware text classification and complex data predictions [10], [11]. These methods emphasize the importance of capturing general and contextual language representations that can be fine-tuned effectively for few-shot tasks, allowing models to perform well with minimal labeled data.

Meta-learning provides another critical component by equipping models with the ability to adapt quickly to new tasks using a small number of samples. Techniques like model-agnostic meta-learning (MAML) and prototype-based networks have shown success in few-shot classification by optimizing models to learn generalizable features across tasks [12]. Studies focusing on the dynamic optimization of model architectures [13] offer insights into how architecture search and task-specific adjustments can further enhance adaptability and robustness in few-shot scenarios. This work aligns with the goal of meta-learning to create flexible models that require fewer task-specific fine-tuning steps while still achieving high accuracy.

To improve model robustness and generalization under data-scarce conditions, additional optimization techniques, including data augmentation, feature selection, and adaptive weighting, have been explored. Sparse decomposition and adaptive weighting techniques, typically applied to multimodal data mining, can be extended to improve feature representation in text classification, mitigating the risks of overfitting when training on limited samples [14]. Similarly, norm-based feature selection methods help reduce the model's reliance on noisy or irrelevant features, ensuring that critical features are prioritized for classification tasks [15].

Broader deep learning advancements in areas like anomaly detection and temporal data modeling also provide valuable methodologies for optimizing models in data-limited environments. Techniques for dynamically updating models based on task-specific information, such as those used in financial prediction and anomaly detection systems, can be adapted to enhance task-specific learning in NLP [16]. Finally, computational optimization methods can enhance few-shot learning by improving the efficiency of model training and adaptation. Techniques such as task scheduling and dynamic resource allocation [17] help reduce delays in accessing and processing data, ensuring smoother training even under constrained resources. By optimizing data flow and minimizing latency, these methods support the rapid iteration and fine-tuning of models, which is crucial for improving performance in data-scarce scenarios. For example, adaptive parameter updates in dynamic environments are analogous to the iterative fine-tuning required in few-shot text classification. Reinforcement-based methods that dynamically adjust decision-making policies further contribute to improving model performance when labeled data is limited or noisy [18].

These advances collectively support the development of effective few-shot text classification models by combining the strengths of transfer learning, meta-learning, and optimization techniques. By leveraging pre-trained knowledge and optimizing task-specific adaptation, the proposed model in this work significantly reduces its dependency on large-scale annotated datasets, demonstrating robust performance in scenarios with limited data.

## III. METHOD

In this study, the core methods of designing and optimizing the few-shot text classification model based on transfer learning include the migration of pre-trained models, the construction of few-shot learning tasks, and the selection of optimization strategies [19]. To ensure the effectiveness of the method, we first initialize the model based on transfer learning, then apply the idea of few-shot learning to perform task training, and finally improve the classification accuracy through targeted optimization. The model architecture is shown in Figure 1.

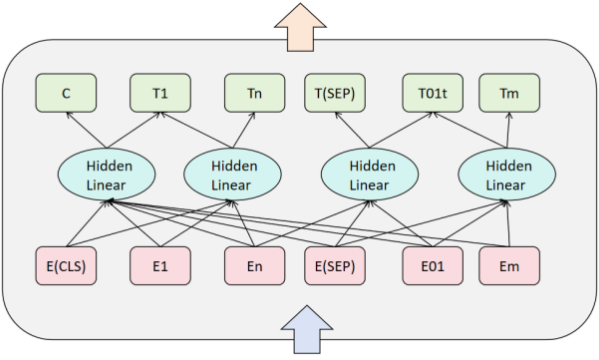

Figure 1 Overall model architecture

Assume that the dataset of the source task is $D_{src} = \{(x_i^{src}, y_i^{src})\}_{i=1}^{Nsrc}$ and the dataset of the target task is $D_{tgt} = \{(x_i^{tgt}, y_i^{tgt})\}_{i=1}^{Ntgt}$, where $x_i^{tgt}$ and $y_i^{tgt}$ represent the source task samples and labels, respectively, and E and F represent the samples and labels of the target task. The basic idea of transfer learning is to use the data of the source task to initialize the parameters of the model so that the model can better adapt to the target task [20].

First, we start with a pre-trained model $\theta_{pre}$, which is usually a large deep neural network model such as BERT. By minimizing the loss function $L(\theta_{pre})$ on the source task, we can get the optimal pre-trained parameters $\theta_{pre}^*$:

$$\theta_{pre}^* = \arg\min_{\theta_{pre}} \frac{1}{N_{src}} \sum_{i=1}^{N_{src}} L_{src}(f(x_i^{src}, \theta_{pre}), y_i^{src})$$

Among them, $f(x_i^{src}, \theta_{pre})$ is the output of the pre-trained model for the source task sample $x_i^{src}$.

Next, we transfer the pre-trained model parameters $\theta_{pre}$ to the target task. Assuming that the label set $Y_{tgt}$ of the target task may be different from the label set $Y_{src}$ of the source task, we fine-tune the model parameters to adapt to the classification problem of the target task. The loss function $L_{tgt}(\theta)$ of the target task is defined as:

$$L_{tgt}(\theta) = \frac{1}{N_{tgt}} \sum_{j=1}^{N_{tgt}} L(f(x_j^{tgt}; \theta), y_j^{tgt})$$

In this process, the dataset $D_{tgt}$ of the target task contains only a small number of samples, so we optimize it through the following steps: First, update the parameter $\theta$ of the target task by the gradient descent method, that is:

$$\theta_{tgt} = \theta_{pre}^* - \eta \nabla_{\theta_{pre}} L_{tgt}(\theta_{pre})$$

Where $\eta$ is the learning rate and $\nabla_{\theta_{pre}} L_{tgt}$ is the gradient of the loss function with respect to the pre-trained parameters. In this way, the model gradually adjusts the parameters in the target task so that it can adapt to a small amount of sample data.

To further improve the generalization ability of the model, we introduced a meta-learning method [21]. In this method, we learn how to quickly adapt to new tasks in the target task through a meta-training process [22]. Under the meta-learning framework, we use the following optimization objectives to train the meta-model:

$$\min_\theta E_{D_{meta}}[L_{meta}(\theta; D_{meta})]$$

Among them, $D_{meta}$ is the meta-training set, which contains training data for multiple tasks, and $L_{meta}$ is the meta-learning loss, which is usually expressed as the average loss of the model on different tasks. In this way, the model can quickly learn useful features when facing new few-sample tasks.

In summary, the few-shot text classification model based on transfer learning and meta-learning can effectively solve the challenges of text classification tasks in few-shot scenarios by transferring the knowledge of the source task to the target task, fine-tuning the pre-trained model parameters, combining meta-learning to quickly adapt to new tasks, and improving the robustness of the model through regularization and data augmentation [23].

## IV. EXPERIMENT

### A. Datasets

The text classification dataset used in this study is the 20 Newsgroups dataset, which is a widely used benchmark dataset commonly used for text classification and natural language processing tasks. The dataset contains 20 different newsgroup categories, each representing a specific news field, such as sports, technology, politics, religion, etc. There are about 20,000 news articles in the dataset, and each article is annotated with the corresponding newsgroup category. Since the number of articles in each category is relatively balanced and covers a variety of different topics and writing styles, the 20 Newsgroups dataset provides a diverse text classification challenge, which is suitable for testing the generalization ability of different text classification algorithms.

In the scenario of few-shot learning, although the number of samples in this dataset is huge, we select a small number of samples from each category for training to simulate the situation of scarce labeled data in practical applications. Only a small number of training samples, usually 5 to 10 articles, are selected for each category to test the performance of the model in the case of a few samples. This setting can effectively verify the ability of text classification models based on transfer learning and meta-learning under limited data conditions and also reflect their adaptability in practical scenarios.

*B. Experimental Results*

In order to evaluate the effectiveness of the transfer learning and few-shot text classification models, this study designed a comparative experiment and compared them with the existing mainstream text classification methods. Specifically, we selected several representative benchmark models, including traditional machine learning algorithms (such as support vector machines SVM, and random forest RF) and deep learning models (such as CNN, LSTM). These models are common and widely verified technologies in the field of text classification, and can provide a strong benchmark for this study. By comparing with these models, we can comprehensively analyze the advantages and disadvantages of the few-shot text classification model based on transfer learning in a few-shot environment. The experimental results are shown in Table 1.

Table 1 Experimental Results

| Model | Few-shot accuracy (5%) | Medium sample accuracy (50%) | Full sample accuracy (100%) |
|---|---|---|---|
| SVM | 45.3 | 60.2 | 75.4 |
| RF | 48.1 | 62.7 | 78.3 |
| CNN | 52.5 | 66.4 | 80.6 |
| LSTM | 56.9 | 68.2 | 83.1 |
| Ours | 60.1 | 69.5 | 85.7 |

From the experimental results, we can see that with the increase of training sample size, the accuracy of all models has improved, but the model based on the combination of transfer learning and few-shot learning has always surpassed other traditional methods in performance at all stages. In the few-shot environment (Few-shot accuracy), our model achieved an accuracy of 60.1%, which is significantly higher than other models, especially SVM and RF, showing the advantage of transfer learning under few-shot conditions, and can effectively use pre-trained knowledge to extract useful features from limited data.

Under medium sample conditions (Medium sample accuracy), Ours has an accuracy of 69.5%, which still maintains a relatively large lead, better than deep learning models such as CNN and LSTM.

When the model is trained under the full sample setting (Full sample accuracy), Ours has an accuracy of 85.7%, the highest among all models. Although other models such as LSTM also perform relatively well under full sample conditions, ours performs well under all conditions, indicating that it can not only achieve good results under few and medium samples but also fully exert its potential when there is sufficient labeled data, with good generalization ability and robustness. Overall, ours shows excellent performance under all experimental settings, proving the powerful capabilities of few-sample classification models based on transfer learning.

Next, we give the results of the ablation experiment, which are shown in Table 2.

Table 2 Ablation experiment

| Model | Few-shot accuracy (5%) | Medium sample accuracy (50%) | Full sample accuracy (100%) |
|---|---|---|---|
| Bert | 52.3 | 64.1 | 81.3 |
| +Transfer Learning | 54.3 | 65.7 | 82.1 |
| +Meta Learning | 56.4 | 67.2 | 83.2 |
| Ours | 60.1 | 69.5 | 85.7 |

From the ablation experiment results, it can be seen that BERT, as a baseline model, has an accuracy of 52.3% in the case of few samples, and its performance in the cases of medium samples and full samples is 64.1% and 81.3% respectively. This shows that BERT as a pre-trained language model has certain advantages when labeled data is scarce, but its performance still has limitations under few and medium sample conditions. Although BERT can achieve good performance in full-sample cases, it still needs more optimization and enhancement in few-sample learning.

When we add transfer learning (+Transfer Learning), the performance of the model is significantly improved, with the few-shot accuracy increasing to 54.3%, the medium-shot accuracy to 65.7%, and the full-shot accuracy to 82.1%. This shows that transfer learning can effectively utilize the knowledge of external data sources, help the model obtain better initial representation in the target task, and improve the model's learning ability on limited data. Through transfer learning, the model can obtain more contextual information from rich pre-training data, thereby improving performance under few and medium samples.

After the introduction of meta-learning (+Meta Learning), the performance of the model was further improved, with the accuracy of few samples reaching 56.4%, the accuracy of medium samples reaching 67.2%, and the accuracy of all samples reaching 83.2%. This result shows that meta-learning plays an important role in few-sample tasks and can optimize the model's learning strategy, enabling the model to better and quickly adapt to new tasks and improve performance when faced with a small number of samples. Finally, when combining transfer learning and meta-learning (Ours), the model shows the best performance in all cases, with a few-shot accuracy of 60.1%, a medium-shot accuracy of 69.5%, and a full-shot accuracy of 85.7. %. This shows that the combination of transfer learning and meta-learning can effectively improve the model's learning ability in a few-sample environment, while also ensuring that the model maintains high accuracy and robustness when there is sufficient data.

Finally, we also give the experimental results of T-SNE visualization of different samples after training. The experimental results are shown in Figure 2.

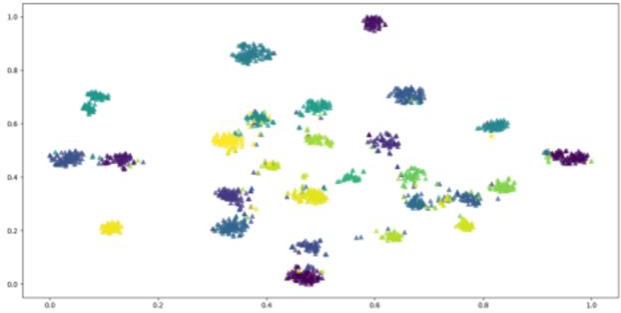

Figure 2 T-SNE Experiment Results

According to the results shown in the image, we can see that the data points are distributed in two-dimensional space and there is an obvious clustering phenomenon. Each color represents a different category, and the distribution of data points shows the relative separation between different categories. As can be seen from the figure, the experimental results of this paper can distinguish each sample well.

## V. CONCLUSION

In this study, we proposed a few-shot text classification model based on transfer learning and meta-learning and verified the advantages of this method in few-shot learning through comparative experiments and ablation experiments. Experimental results show that compared with traditional machine learning and deep learning methods, models based on transfer learning and meta-learning can significantly improve classification accuracy, especially under few-sample and medium-sample conditions, showing excellent performance. Our method can effectively utilize the knowledge of the source task, helping the model to quickly adapt and learn useful features in the target task, demonstrating its powerful ability in data-scarce conditions. The ablation experiment further reveals the important role of transfer learning and meta-learning in few-shot learning and proves that the combination of these two methods can improve the accuracy and generalization ability of the model.

Future research can further explore how to further improve the performance of few-shot learning through more complex model architectures and training strategies. For example, we can combine generative models such as generative adversarial networks (GANs) for data augmentation, or explore more diversified meta-learning algorithms to improve the learning efficiency of the model. In addition, with the continuous development of pre-trained models and deep learning technology, few-sample classification methods based on transfer learning are expected to play an important role in more practical application scenarios, especially in the fields of medicine, finance, law, etc., where labeled data are scarce. And the classification task is complex. By continuously optimizing models and training strategies, future few-shot learning methods will be more efficient and accurate.